\documentclass[runningheads]{llncs}
\usepackage[linesnumbered,ruled]{algorithm2e}
\usepackage{times,hyperref,amssymb,amsmath,amsfonts,graphicx,float,subcaption,geometry,latexsym,url,multicol}

\begin{document}
	
\title{Recurrent Auto-Encoder Model for\\ Large-Scale Industrial Sensor Signal Analysis
	\thanks{Supported by Centrica plc. Registered office: Millstream, Maidenhead Road, Windsor SL4 5GD, United Kingdom.}}

\author{Timothy Wong\orcidID{0000-0003-1943-6448} \and \\
Zhiyuan Luo\orcidID{0000-0002-3336-3751}}

\authorrunning{Wong and Luo.}

\institute{
	Royal Holloway, University of London, \\Egham TW20 0EX, \\United Kingdom.
}

\maketitle

\begin{abstract}

Recurrent auto-encoder model summarises sequential data through an encoder structure into a fixed-length vector and then reconstructs the original sequence through the decoder structure. The summarised vector can be used to represent time series features. In this paper, we propose relaxing the dimensionality of the decoder output so that it performs partial reconstruction. The fixed-length vector therefore represents features in the selected dimensions only. In addition, we propose using rolling fixed window approach to generate training samples from unbounded time series data. The change of time series features over time can be summarised as a smooth trajectory path. The fixed-length vectors are further analysed using additional visualisation and unsupervised clustering techniques. The proposed method can be applied in large-scale industrial processes for sensors signal analysis purpose, where clusters of the vector representations can reflect the operating states of the industrial system.

\end{abstract}

\section{Background}

\begin{figure}[H]
	\centering
	\includegraphics[width=0.8\textwidth]{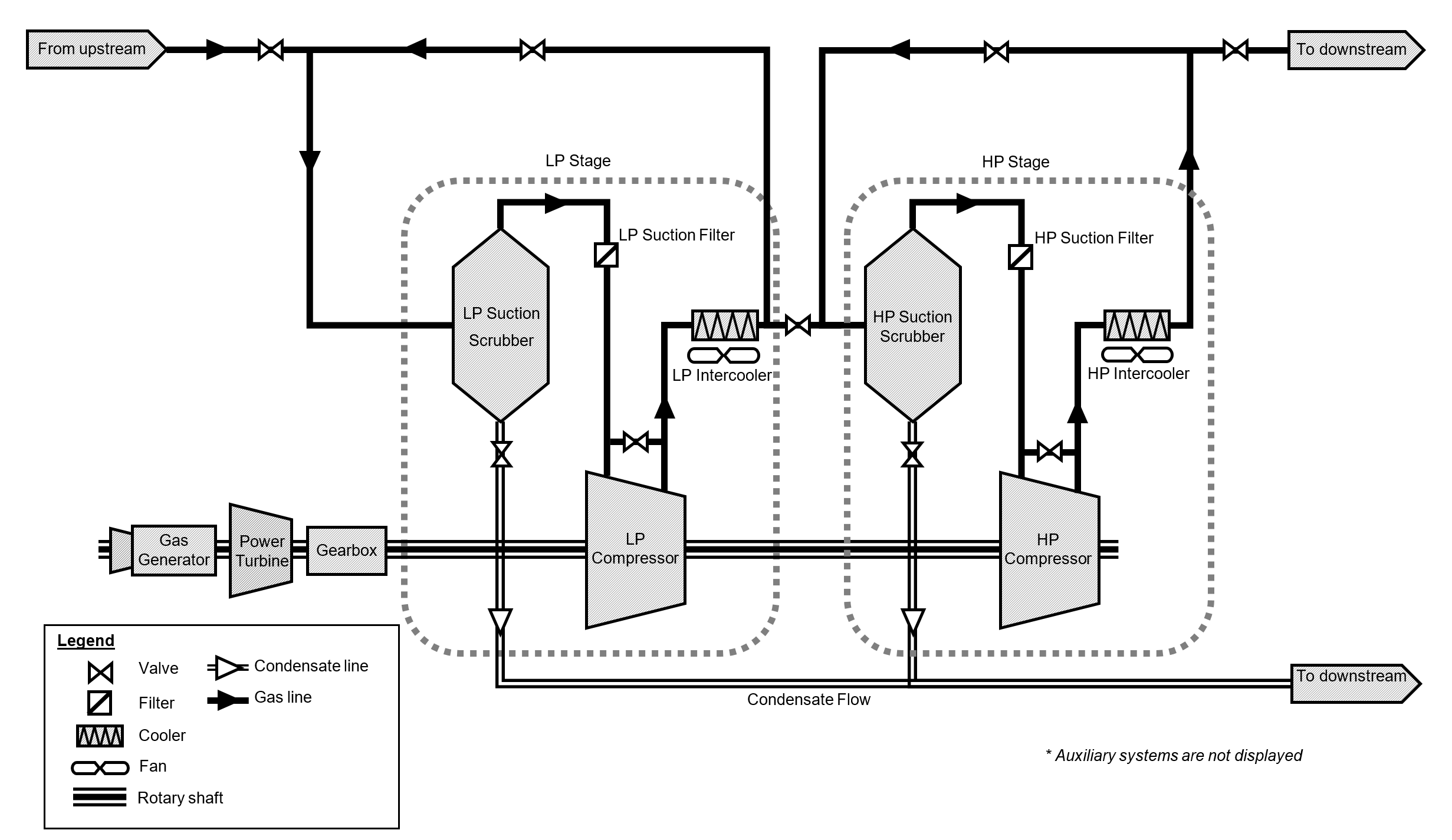}
	\caption{A simplified process diagram of the two-stage centrifugal compression train which is located at a natural gas terminal.}
	\label{fig:process_diagram}
\end{figure}

Modern industrial processes are often monitored by a large array of sensors. Machine learning techniques can be used to analyse unbounded streams of sensor signal in an on-line scenario. This paper illustrates the idea using propietary data collected from a two-stage centrifugal compression train driven by an aeroderivative industrial engine (Rolls-Royce RB211) on a single shaft. This large-scale compression module belongs to a major natural gas terminal\footnote{A simplified process diagram of the compression train can be found in Figure \ref{fig:process_diagram} at the appendix.}. The purpose of this modular process is to regulate the pressure of natural gas at an elevated, pre-set level. At the compression system, sensors are installed to monitor the production process. Real-valued measurements such as temperature, pressure, rotary speed, vibration... etc., are recorded at different locations \footnote{A list of sensors is available in the appendix.}.

Streams of sensor signals can be treated as a multidimensional entity changing through time. Each stream of sensor measurement is basically a set of real values received in a time-ordered fashion. When this concept is extended to a process with \(P\) sensors, the dataset can therefore be expressed as a time-ordered multidimensional vector \( \{ \mathbb{R}_t^P:t\in [1,T] \} \).

The dataset used in this study is unbounded (i.e. continuous streaming) and unlabelled, where the events of interest (e.g. overheating, mechanical failure, blocked oil filters... etc) are not present. The key goal of this study is to identify sensor patterns and anomalies to assist equiptment maintenance. This can be achieved by finding the representation of multiple sensor data. We propose using recurrent auto-encoder model to extract vector representation for multidimensional time series data. Vectors can be analysed further using visualisation and clustering techniques in order to identify patterns.

\subsection{Related Works}

A comprehensive review \cite{bakeoff} analysed traditional clustering algorithms for unidimensional time series data. It has concluded that Dynamic Time Warping (DTW) can be an effective benchmark for unidimensional time series data representation. There has been attempts to generalise DTW to multidimensional level \cite{vlachos,gillian,holt,ko,petitjean,liu,wang,Shokoohi,giorgino}. Most of these studies focused on analysing time series data with relatively low dimensionality, such as those collected from Internet of Things (IoT) devices, wearable sensors and gesture recognition. This paper contributes further by featuring a time series dataset with much higher dimensionality which is representative for any large-scale industrial applications.

Among neural network researches, \cite{srivastava-et-al} proposed a recurrent auto-encoder model based on LSTM neurons which aims at learning video data representation. It achieves this by reconstructing sequence of video frames. Their model was able to derive meaningful representations for video clips and the reconstructed outputs demonstrate sufficient similarity based on qualitative examination. Another recent paper \cite{davino} also used LSTM-based recurrent auto-encoder model for video data representation. Sequence of frames feed into the model so that it learns the intrinsic representation of the underlying video source. Areas with high reconstruction error indicate divergence from the known source and hence can be used as a video forgery detection mechanism.

Similarly, audio clips can treated as sequential data. A study \cite{chung} converted variable-length audio data into fixed-length vector representation using recurrent auto-encoder model. It found that audio segments that sound alike usually have vector representations in same neighbourhood.

There are other works related to time series data. For instance, a recent paper \cite{malhotra} proposed a recurrent auto-encoder model which aims at providing fixed-length representation for bounded univariate time series data. The model was trained on a plurality of labelled datasets with the aim of becoming a generic time series feature extractor. Dimensionality reduction of the vector representation via t-SNE shows that the ground  labels can be observed in the extracted representations. Another study \cite{hsu} proposed a time series compression algorithm using a pair of RNN encoder-decoder structure and an additional auto-encoder to achieve higher compression ratio. Meanwhile, another research \cite{lee-d} used an auto-encoder model with database metrics (e.g. CPU usage, number of active sessions... etc) to identify anomalous usage periods by setting threshold on the reconstruction error.

\section{Methods}

A pair of RNN encoder-decoder structure can provide end-to-end mapping between an ordered multidimensional input sequence and its matching output sequence \cite{sutskever2014,cho2014}. Recurrent auto-encoder can be depicted as a special case of the aforementioned model, where input and output sequences are aligned with each other. It can be extended to the area of signal analysis in order to leverage recurrent neuron’s power to understand complex and time-dependent relationship.

\subsection{Encoder-Decoder Structure} 
At high level, the RNN encoder reads an input sequence and summarises all information into a fixed-length vector. The decoder then reads the vector and reconstructs the original sequence. Figure \ref{fig:seq2seq} below illustrates the model.

\begin{figure}[H]
	\centering
	\includegraphics[width=1\textwidth]{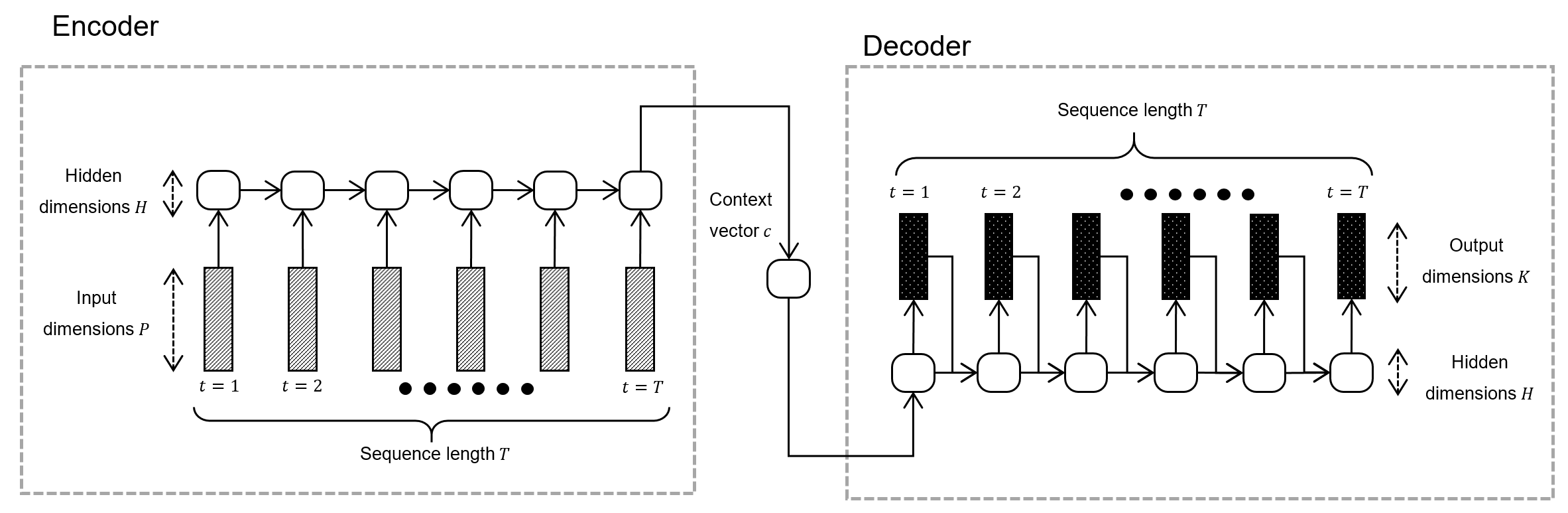}
	\caption{Recurrent auto-encoder model. Both the encoder and decoder are made up of multilayered RNN. Arrows indicate the direction of information flow.}
	\label{fig:seq2seq}
\end{figure}

\subsubsection{Encoding}

The role of the recurrent encoder is to project the multidimensional input sequence into a fixed-length hidden context vector \(c\). It reads the input vectors \(\{\mathbb{R}_t^P:t\in [1,T]\}\) sequentially from \(t=1,2,3,...,T\). The hidden state of the RNN has \(H\) dimensions which updates at every time step based on the current input and hidden state inherited from previous steps.

Recurrent neurons arranged in multiple layers are capable of learning complex temporal behaviours. In this proposed model, LSTM neurons with hyperbolic tangent activation are used at all recurrent layers \cite{hochreiter1997}. An alternative choice of using gated recurrent unit (GRU) neurons \cite{cho2014} can also be used but was not experimented within the scope of this study. Once the encoder reads all the input information, the sequence is summarised in a fixed-length vector \(c\) which has \(H\) hidden dimensions.

For regularisation purpose, dropout can be applied to avoid overfitting. It refers to randomly removing a fraction of neurons during training, which aims at making the network more generalisable \cite{srivastava2014}. In an RNN setting, \cite{zaremba2014} suggested that dropout should only be applied non-recurrent connections. This helps the recurrent neurons to retain memory through time while still allowing the non-recurrent connections to benefit from regularisation.

\subsubsection{Decoding}
The decoder is a recurrent network which uses the representation \(c\) to reconstruct the original sequence. To exemplify this, the decoder starts by reading the context vector \(c\) at \(t=1\). It then decodes the information through the RNN structure and outputs a sequence of vectors \( \{ \mathbb{R}_t^K:t\in [1,T] \} \) where \(K\) denotes the dimensionality of the output sequence. 

Recalling one of the fundamental characteristics of an auto-encoder is the ability to reconstruct the input data back into itself via a pair of encoder-decoder structure. This criterion can be slightly relaxed such that \(K \leqslant P\), which means the output sequence is only a partial reconstruction of the input sequence.

Recurrent auto-encoder with partial reconstruction:
\begin{equation}
\label{seq2seq_autoencoder_relax_encoder}
\begin{cases} 
f_{encoder} : \{ \mathbb{R}_t^P:t \in [1, T] \} \rightarrow c \\
f_{decoder} : c \rightarrow \{ \mathbb{R}_t^K:t \in [1, T] \} \\
\end{cases} K \leqslant P
\end{equation}

In the large-scale industrial system use case, all streams of sensor measurements are included in the input dimensions while only a subset of sensors is included in the output dimensions. This means that the entire system is visible to the encoder, but the decoder only needs to perform partial reconstruction of it. End-to-end training of the relaxed auto-encoder implies that the context vector would summarise the input sequence while still being conditioned on the output sequence. Given that activation of the context vector is conditional on the decoder output, this approach allows the encoder to capture lead variables across the entire process as long as they are relevant to the selected output dimensions. 

It is important to recognise that reconstructing part of the data is an easier task to perform than fully-reconstructing the entire original sequence. However, partial reconstruction has practical significance for industrial applications. In real-life scenarios, multiple context vectors can be generated from different recurrent auto-encoder models using identical sensors in the encoder input but different subset of sensors in the decoder output. The selected subsets of sensors can reflect the underlying operating states of different parts of the industrial system. As a result, context vectors produced from the same temporal segment can be used as different diagnostic measurements in industrial context. We will illustrate this in the results section by two examples.

\subsection{Sampling}

For a training dataset of \(T^\prime\) time steps, samples can be generated where \(T < T^\prime\). We can begin at \(t=1\) and draw a sample of length \(T\). This process continues recursively by shifting one time step until it reaches the end of the training dataset. For a subset sequence with length \(T\), this method allows \(T^\prime - T\) samples to be generated. Besides, it can also generate samples from an unbounded time series in an on-line scenrio, which are essential for time-critical applications such as sensor data analysis.

\begin{algorithm}[H]
	\label{consecutive_sampling}
	\caption{Drawing samples consecutively from the original dataset}
	\SetKwInOut{Input}{Input}
	\Input{Dataset length \(T^\prime\)}
	\Input{Sample length \(T\)}
	\(i\leftarrow 0\) \;
	\While{\(i \leqslant i+T \) } {
		Generate sample sequence \((i, i+T]\) from the dataset\;
		\(i\leftarrow i+1\)\;
	}
\end{algorithm}

Given that sample sequences are recursively generated by shifting the window by one time step, successively-generated sequences are highly correlated with each other. As we have discussed previously, the RNN encoder structure compresses sequential data into a fixed-length vector representation. This means that when consecutive sequences are fed through the encoder structure, the resulting activation at \(c\) would also be highly correlated. As a result, consecutive context vectors can join up to form a smooth trajectory in space.

Context vectors in the same neighbourhood have similar activation therefore the industrial system must have similar underlying operating states. Contrarily, context vectors located in distant neighbourhoods would have different underlying operating states. These context vectors can be visualised in lower dimensions via dimensionality reduction techniques such as principal component analysis (PCA).

Furthermore, additional unsupervised clustering algorithms can be applied to the context vectors. Each context vector can be assigned to a cluster \(C_j\) where \(J\) is the total number of clusters. Once all the context vectors are labelled with their corresponding clusters, supervised classification algorithms can be used to learn the relationship between them using the training set. For instance, support vector machine (SVM) classifier with \(J\) classes can be used. The trained classifier can then be applied to the context vectors in the held-out validation set for cluster assignment. It can also be applied to context vectors generated from unbounded time series in an on-line setting. Change in cluster assignment among successive context vectors indicates a change in the underlying operating state.

\section{Results}

Training samples were drawn from the dataset using windowing approach with fixed sequence length. In our example, the large-scale industrial system has \(158\) sensors which means the recurrent auto-encoder's input dimension has \(P=158\). Observations are taken at \(5\) minutes granularity and the total duration of each sequence was set at \(3\) hours. This means that the model's sequence has fixed length \(T=36\), while samples were drawn from the dataset with total length \(T^\prime=2724\). The dataset was scaled into \(z\)-scores, thus ensuring zero-centred data which facilitates gradient-based training.

The recurrent auto-encoder model has three layers in the RNN encoder structure and another three layers in the corresponding RNN decoder. There are \(400\) neurons in each layer. The auto-encoder model structure can be summarised as: RNN encoder (\(400\) neurons / \(3\) layers LSTM / hyperbolic tangent) - Context layer (\(400\) neurons / Dense / linear activation) - RNN decoder (\(400\) neurons / \(3\) layers LSTM / hyperbolic tangent). Adam optimiser \cite{kingma} with \(0.4\) dropout rate was used for model training.

\subsection{Output Dimensionity}

As we discussed earlier, the RNN decoder's output dimension can be relaxed for partial reconstruction. The output dimensionality was set at \(K=6\) which is comprised of a selected set of sensors relating to key pressure measurements (e.g. suction and discharge pressures of the compressor device).

We have experimented three scenarios where the first two have complete dimensionality \(P=158; K=158\) and \(P=6; K=6\) while the remaining scenario has relaxed dimensionality \(P=158; K=6\). The training and validation MSEs of these models are visualised in figure \ref{fig:output_dims} below.

\begin{figure}[h]
	\centering
	\includegraphics[width=0.65\textwidth]{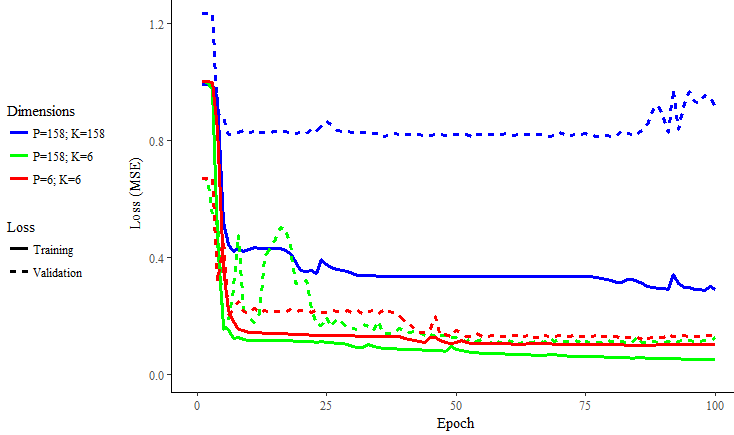}
	\caption{Effects of relaxing dimensionality of the output sequence on the training and validation MSE losses. They contain same number of layers in the RNN encoder and decoder respectively. All hidden layers contain same number of LSTM neurons with hyperbolic tangent activation.}
	\label{fig:output_dims}
\end{figure}

The first model with complete dimensionality (\(P=158; K=158\)) has visibility of all dimensions in both the encoder and decoder structures. Yet, both the training and validation MSEs are high as the model struggles to compress-decompress the high dimensional time series data.

For the complete dimensionality model with \(P=6; K=6\), the model has limited visibility to the system as only the selected dimensions were included. Despite the context layer summarises information specific to the selected dimensionality in this case, lead variables in the original dimensions have been excluded. This prevents the model from learning any dependent behaviours among all available information.

On the other hand, the model with partial reconstruction (\(P=158; K=6\)) demonstrate substantially lower training and validation MSEs. Since all information is available to the model via the RNN encoder, it captures the relevant information such as lead variables across the entire system. 

Randomly selected samples in the held-out validation set were fed to this model and the predictions can be qualitatively examined in details. In figure~\ref{fig:heatmaps} below, all the selected specimens demonstrate high similarity between the original label and the reconstructed output. The recurrent auto-encoder model captures the shift in mean level as well as temporal variations across all output dimensions. 

\begin{figure}[H]
	\centering
	\includegraphics[width=.8\textwidth]{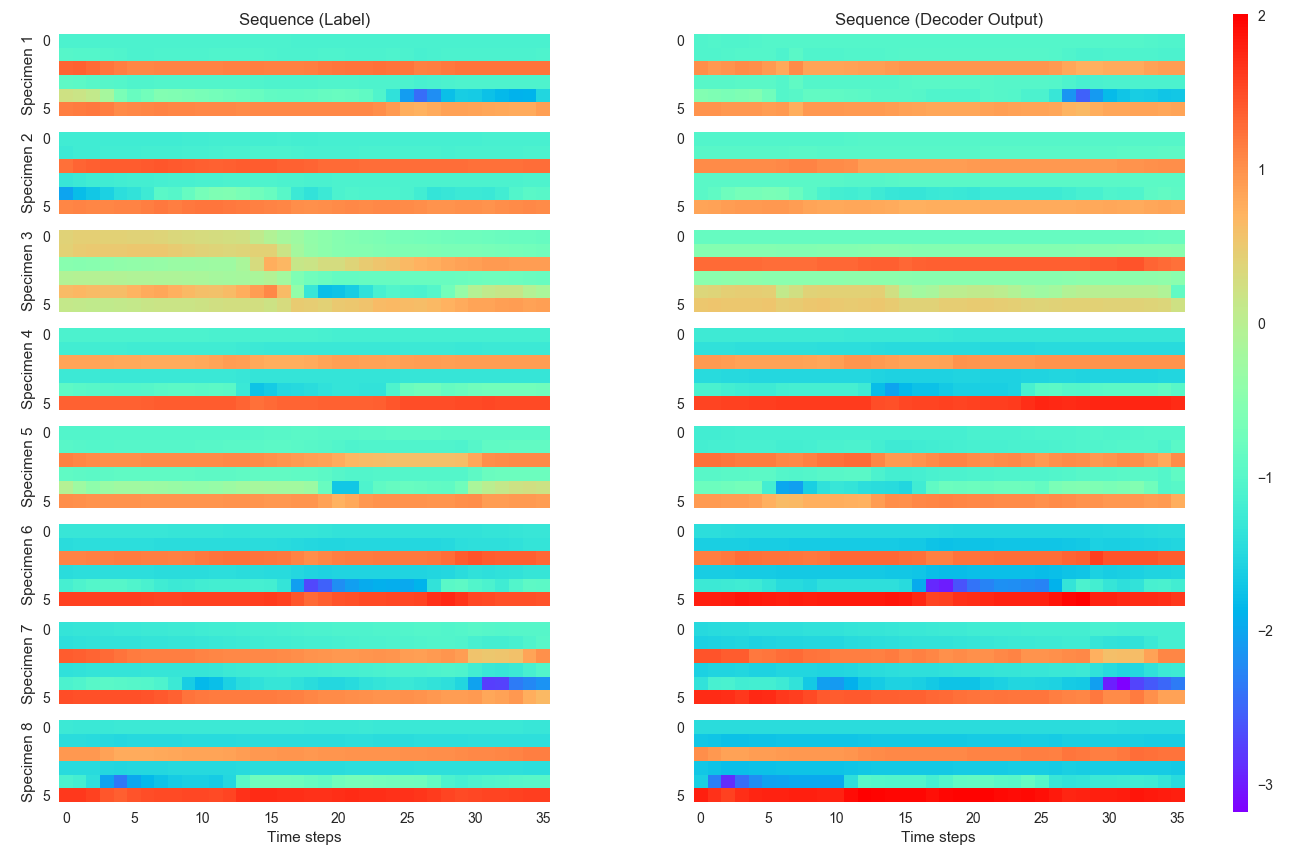}
	\caption{A heatmap showing eight randomly selected output sequences in the held-out validation set. Colour represents magnitude of sensor measurements in normalised scale.}
	\label{fig:heatmaps}
\end{figure}

\subsection{Context Vector}

Once the recurrent auto-encoder model is successfully trained, samples can be fed to the model and the corresponding context vectors can be extracted for detailed inspection. In the model we selected, the context vector \(c\) is a multi-dimensional real vector \(\mathbb{R}^{400}\). Since the model has input dimensions \(P=158\)  and sequence length \(T=36\), the model has achieved compression ratio \(\frac{158\times36}{400}=14.22\). Dimensionality reduction of the context vectors through principal component analysis (PCA) shows that context vectors can be efficiently embedded in lower dimensions (e.g. two-dimensional space).

At low-dimensional space, we used supervised classification algorithm to learn the relationship between vectors representations and cluster assignment. The trained classification model can then be applied to the validation set to assign clusters for unseen data. In our experiment, a SVM classifier with radial basis function (RBF) kernel (\(\gamma=4\)) was used. The results are shown in figure \ref{fig:pca_cluster} below.

\begin{figure}[H]
	\centering
	\begin{subfigure}[b]{0.5\textwidth}
		\includegraphics[width=\textwidth]{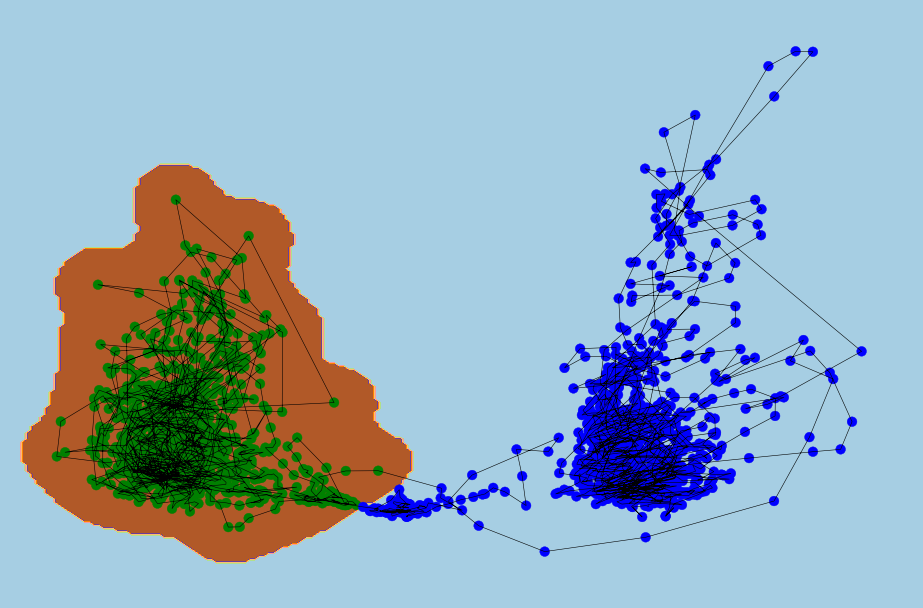}
		\caption{\(2\) clusters}
		\label{fig:pca_cluster_2}
	\end{subfigure}
	~
	\begin{subfigure}[b]{0.25\textwidth}
		\includegraphics[width=\textwidth]{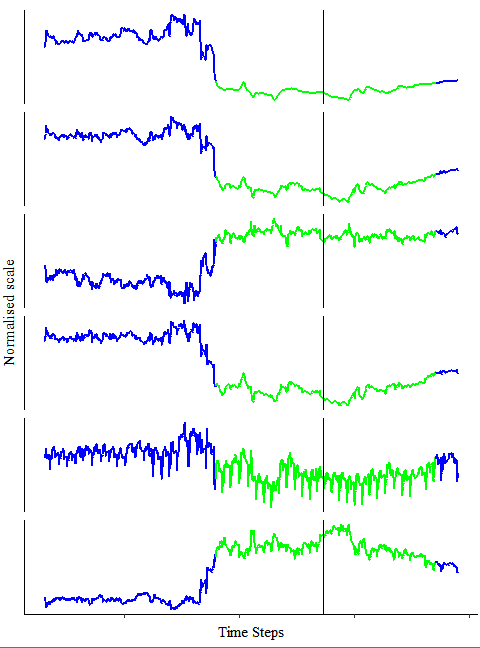}
		\label{fig:context_timeline_2}
	\end{subfigure}
	
	\begin{subfigure}[b]{0.5\textwidth}
		\includegraphics[width=\textwidth]{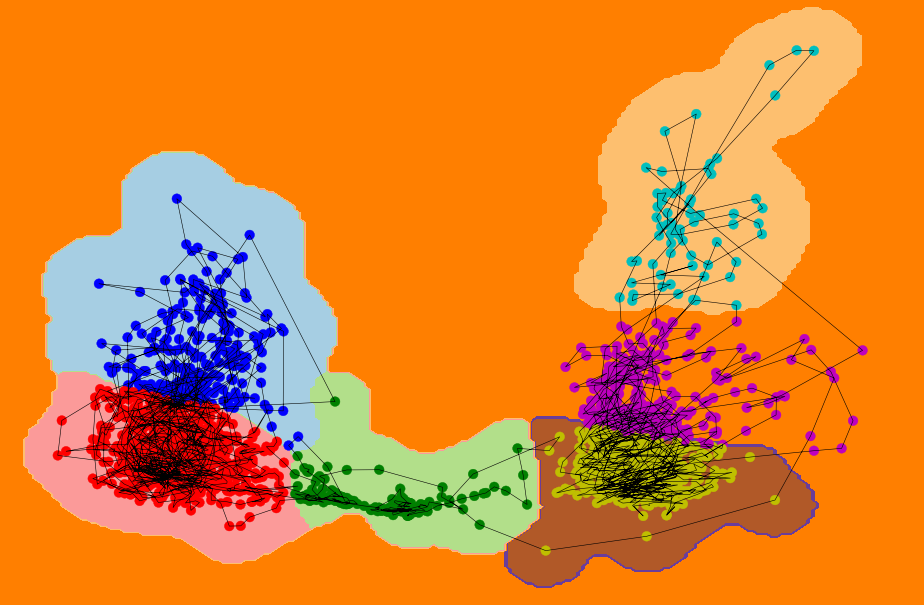}
		\caption{\(6\) clusters}
		\label{fig:pca_cluster_6}
	\end{subfigure}
	~
	\begin{subfigure}[b]{0.25\textwidth}
		\includegraphics[width=\textwidth]{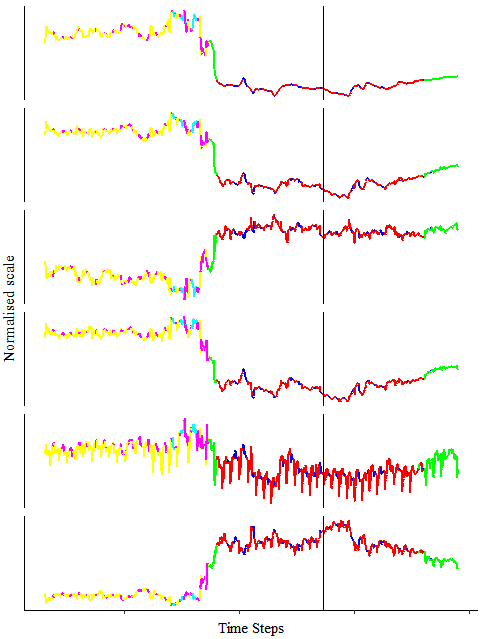}
		\label{fig:context_timeline_6}
	\end{subfigure}
	
	\caption{The first example. On the left, the context vectors were projected into two-dimensional space using PCA. The black solid line on the left joins all consecutive context vectors together as a trajectory. Different number of clusters were identified using simple \(K\)-means algorithm. Cluster assignment and the SVM decision boundaries are coloured in the charts. On the right, output dimensions are visualised on a shared time axis. The black solid line demarcates the training set (\(70\%\)) and validation sets (\(30\%\)). The line segments are colour-coded to match the corresponding clusters.}
	\label{fig:pca_cluster}
\end{figure}

In two-dimensional space, the context vectors separate into two clearly identifiable neighbourhoods. These two distinct neighbourhoods correspond to the shift in mean values across all output dimensions. When \(K\)-means clustering algorithm is applied, it captures these two neighbourhoods as two clusters in the scenario depicted in figure~\ref{fig:pca_cluster_2}.

When the number of clusters increases, they begin to capture more subtleties. In the six clusters scenario illustrated in figure~\ref{fig:pca_cluster_6}, successive context vectors oscillate back and forth between neighbouring clusters. The trajectory corresponds to the interlacing troughs and crests in the output dimensions. Similar pattern can also be observed in the validation set, which indicates that the knowledge learned by the auto-encoder model is generalisable to unseen data.

\begin{figure}[H]
	\centering
	\begin{subfigure}[b]{0.35\textwidth}
		\includegraphics[width=\textwidth]{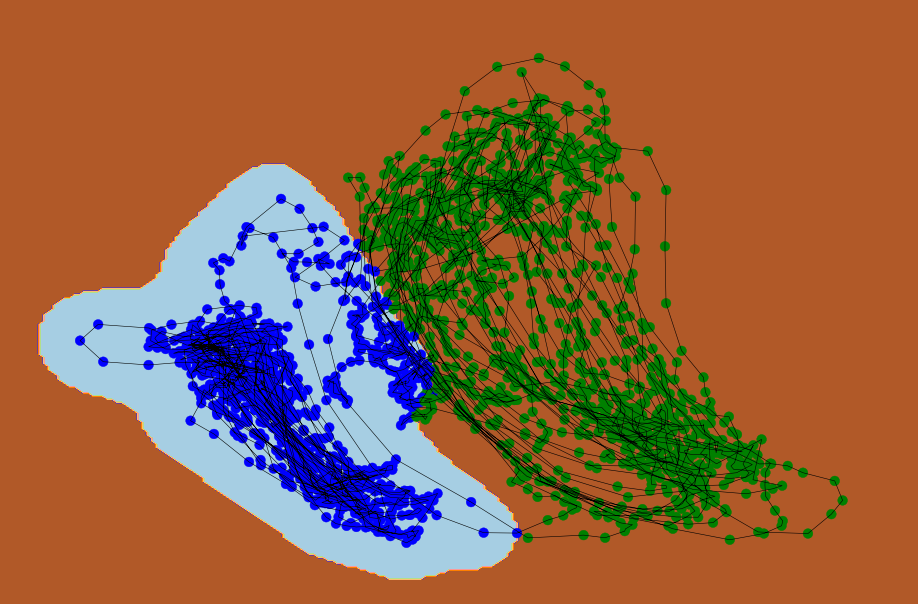}
		\caption{\(2\) clusters}
		\label{fig:ex2_pca_cluster_2}
	\end{subfigure}
	~
	\begin{subfigure}[b]{0.6\textwidth}
		\includegraphics[width=\textwidth]{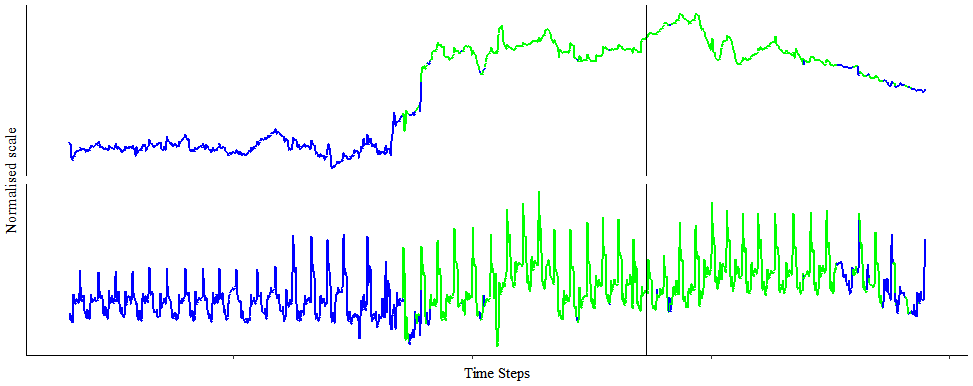}
		\label{fig:ex2_context_timeline_2}
	\end{subfigure}
	
	\begin{subfigure}[b]{0.35\textwidth}
		\includegraphics[width=\textwidth]{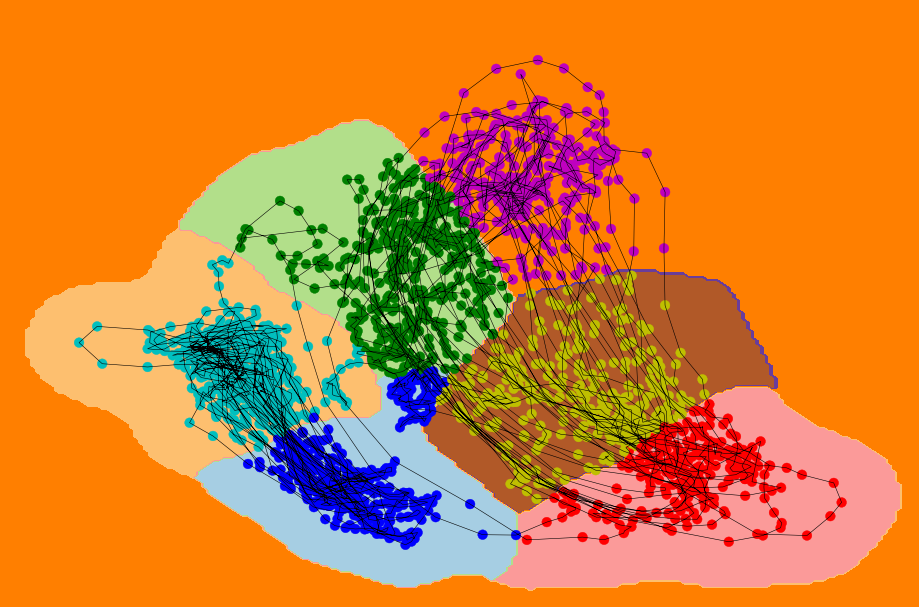}
		\caption{\(6\) clusters}
		\label{fig:ex2_pca_cluster_6}
	\end{subfigure}
	~
	\begin{subfigure}[b]{0.6\textwidth}
		\includegraphics[width=\textwidth]{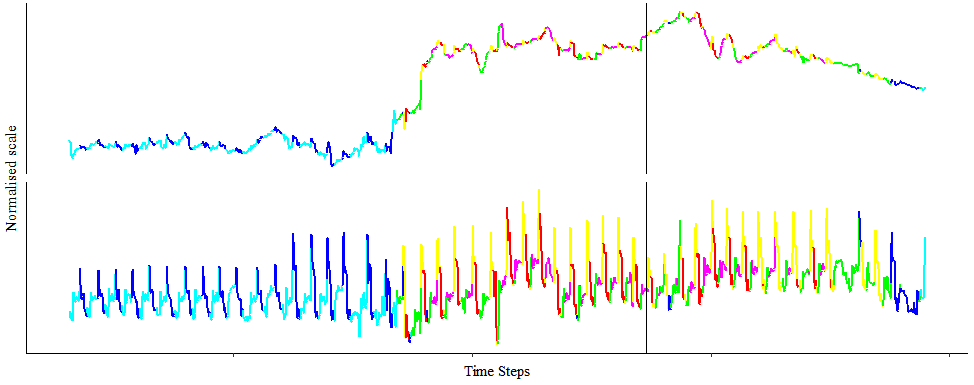}
		\label{fig:ex2_context_timeline_6}
	\end{subfigure}
	
	\caption{The second example. The sensor data is drawn from the same time period as the previous example, only the output dimension has been changed to \(K=2\) where another set of gas pressure sensors were selected. }
	\label{fig:ex2_pca_cluster}
\end{figure}

Furthermore, we have repeated the same experiment again with a different configuration (\(K=158; P=2\)) to reassure that the proposed approach can provide robust representations of the data. The sensor measurements are drawn from an identical time period and only the output dimensionality \(K\) is changed (The newly selected set of sensors is comprised of a different measurements of discharge gas pressure at the compressor unit). Through changing the output dimensionality \(K\), we can illustrate the effects of partial reconstruction using different output dimensions. As seen in figure \ref{fig:ex2_pca_cluster}, the context vectors form a smooth trajectory in the low-dimensional space. Similar sequences yield context vectors which are located in a shared neighbourhood. Nevertheless, the clusters found by \(K\)-means method in this secondary example also manage to identify neighbourhoods with similar sensor patterns.

\section{Discussion and Conclusion}

Successive context vectors generated by windowing approach are always highly correlated, thus form a smooth trajectory in high-dimensional space. Additional dimensionality reduction techniques can be applied to visualise the change of time series features. One of the key contributions of this study is that similar context vectors can be grouped into clusters using unsupervised clustering algorithms such as \(K\)-means algorithm. Clusters can be optionally labelled manually to identify operating state (e.g. healthy vs. faulty). Alarm can be triggered when the context vector travels beyond the boundary of a predefined neighbourhood. Clusters of the vector representation can be used by operators and engineers to aid diagnostics and maintenance.

Another contribution of this study is that dimensionality of the output sequence can be relaxed. This allows the recurrent auto-encoder to perform partial reconstruction. Although it is easier for the model to reconstruct part of the original sequence, such simple improvement allows users to define different sets of sensors of particular interest. By changing sensors in the decoder output, context vectors can be used to reflect underlying operating states of various aspects of the large-scale industrial process. This ultimately enables users to diagnose the industrial system by generating more useful insight.

This proposed method essentially performs multidimensional time series clustering. We have demonstrated that it can natively scale up to very high dimensionality as it is based on recurrent auto-encoder model. We have applied the method to an industrial sensor dataset with \(P=158\) and empirically show that it can represent multidimensional time series data effectively. In general, this method can be further generalised to any multi-sensor multi-state processes for operating state recognition. 

This study established that recurrent auto-encoder model can be used to analyse unlabelled and unbounded time series data. It further demontrated that operating state (i.e. labels) can be inferred from unlabelled time series data. This opens up further possibilities for analysing complex industrial sensors data given that it is predominately overwhelmed with unbounded and unlabelled time series data.

Nevertheless, the proposed approach has not included any categorical sensor measurements (e.g. open/closed, tripped/healthy, start/stop... etc). Future research can focus on incorporating categorical measurements alongside real-valued measurements.

\subsection*{Disclosure}

The technical method described in this paper is the subject of British patent application GB1717651.2.

\bibliographystyle{splncs04}
\bibliography{mybibliography}

\newpage
\section*{Appendix A}

The rotary components are driven by industrial RB-211 jet turbine on a single shaft through a gearbox. Incoming natural gas passes through the low pressure (LP) stage first which brings it to an intermediate pressure level, it then passes through the high pressure (HP) stage and reaches the pre-set desired pressure level. The purpose of the suction scrubber is to remove any remaining condensate from the gas prior to feeding through the centrifugal compressors. Once the hot compressed gas is discharged from the compressor, its temperature is lowered via the intercoolers.

\begin{figure}[H]
	\centering
	\includegraphics[width=0.5\textwidth]{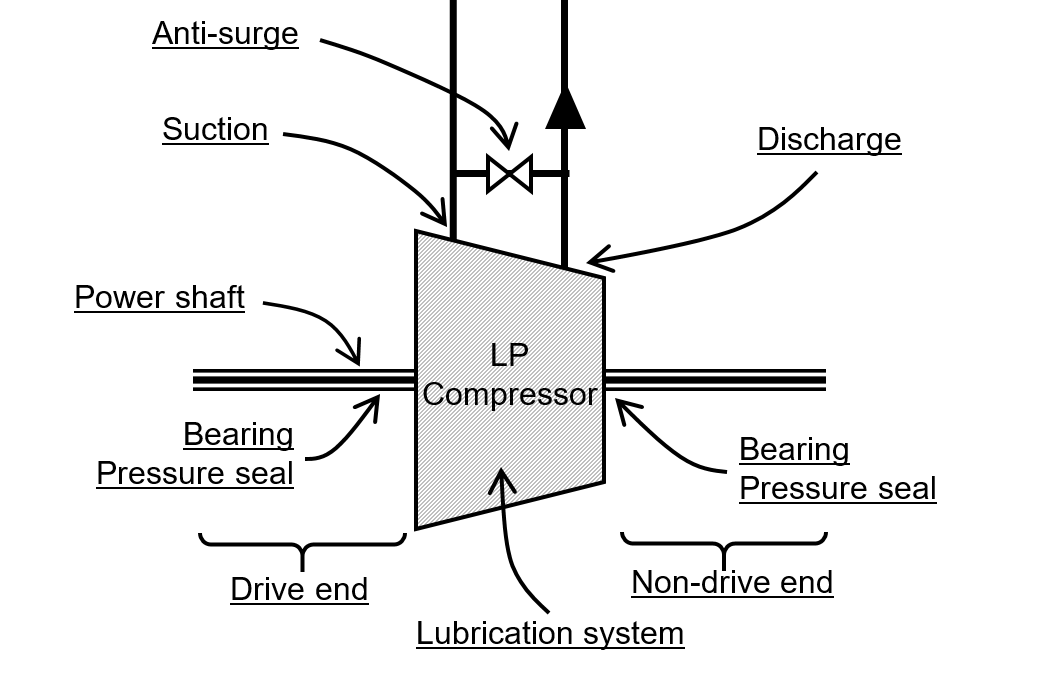}
	\caption{Locations of key components around the centrifugal compressor.}
	\label{fig:lp_stage}
\end{figure}

\newpage
\section*{Appendix B}

The sensor measurements used in the analysis are listed below:

\begin{multicols}{3}
	\begin{enumerate}
		\scriptsize{
			\item GASCOMPCARBONDIOXIDEMEAS
			\item GASCOMPMETHANEMEAS
			\item GASCOMPNITROGENMEAS
			\item GASPROPMOLWTMEAS
			\item PRESSAMBIENT
			\item GB\_SPEEDINPUT
			\item GB\_SPEEDOUTPUT
			\item GB\_TEMPINPUTBRGDRIVEEND
			\item GB\_TEMPINPUTBRGNONDRIVEEND
			\item GB\_TEMPINPUTBRGTHRUSTINBOARD
			\item GB\_TEMPINPUTBRGTHRUSTOUTBRD
			\item GB\_TEMPLUBOIL
			\item GB\_TEMPLUBOILTANK
			\item GB\_TEMPOUTPUTBRGDRIVEEND
			\item GB\_TEMPOUTPUTBRGNONDRIVEEND
			\item GB\_VIBBRGCASINGVEL
			\item GB\_VIBINPUTAXIALDISP
			\item GB\_VIBINPUTDRIVEEND
			\item GB\_VIBINPUTNONDRIVEEND
			\item GB\_VIBOUTPUTDRIVEEND
			\item GB\_VIBOUTPUTNONDRIVEEND
			\item GG\_FLOWFUEL
			\item GG\_FLOWWATERINJECTION
			\item GG\_FLOWWATERINJSETPOINT
			\item GG\_POWERSHAFT
			\item GG\_PRESSAIRINLET
			\item GG\_PRESSCOMPDEL
			\item GG\_PRESSCOMPDELHP
			\item GG\_PRESSCOMPDELIP
			\item GG\_PRESSDIFBRGLUBOIL
			\item GG\_PRESSDIFINLETFILTER
			\item GG\_PRESSDIFINLETFLARE
			\item GG\_PRESSDIFVALVEWATERINJCTRL
			\item GG\_PRESSDISCHWATERINJPUMP1
			\item GG\_PRESSDISCHWATERINJPUMP2
			\item GG\_PRESSEXH
			\item GG\_PRESSFUELGAS
			\item GG\_PRESSHYDOILDEL
			\item GG\_PRESSLUBEOILHEADER
			\item GG\_PRESSLUBOIL
			\item GG\_PRESSMANIFOLDWATERINJ
			\item GG\_PRESSSUCTWATERINJPUMP
			\item GG\_SPEEDHP
			\item GG\_SPEEDIP
			\item GG\_TEMPAIRINLET
			\item GG\_TEMPCOMPDEL
			\item GG\_TEMPCOMPDELHP
			\item GG\_TEMPCOMPDELIP
			\item GG\_TEMPEXH
			\item GG\_TEMPEXHTC1
			\item GG\_TEMPEXHTC2
			\item GG\_TEMPEXHTC3
			\item GG\_TEMPEXHTC4
			\item GG\_TEMPEXHTC5
			\item GG\_TEMPEXHTC6
			\item GG\_TEMPEXHTC7
			\item GG\_TEMPEXHTC8
			\item GG\_TEMPFUELGAS
			\item GG\_TEMPFUELGASG1
			\item GG\_TEMPFUELGASLINE
			\item GG\_TEMPHSOILCOOLANTRETURN
			\item GG\_TEMPHSOILMAINRETURN
			\item GG\_TEMPLUBOIL
			\item GG\_TEMPLUBOILTANK
			\item GG\_TEMPPURGEMUFF
			\item GG\_TEMPWATERINJSUPPLY
			\item GG\_VALVEWATERINJECTCONTROL
			\item GG\_VANEINLETGUIDEANGLE
			\item GG\_VANEINLETGUIDEANGLE1
			\item GG\_VANEINLETGUIDEANGLE2
			\item GG\_VIBCENTREBRG
			\item GG\_VIBFRONTBRG
			\item GG\_VIBREARBRG
			\item HP\_HEADANTISURGE
			\item HP\_POWERSHAFT
			\item HP\_PRESSCLEANGAS
			\item HP\_PRESSDIFANTISURGE
			\item HP\_PRESSDIFSUCTSTRAINER
			\item HP\_PRESSDISCH
			\item HP\_PRESSSEALDRYGAS
			\item HP\_PRESSSEALLEAKPRIMARYDE1
			\item HP\_PRESSSEALLEAKPRIMARYDE2
			\item HP\_PRESSSEALLEAKPRIMARYNDE1
			\item HP\_PRESSSEALLEAKPRIMARYNDE2
			\item HP\_PRESSSUCT1
			\item HP\_PRESSSUCT2
			\item HP\_SPEED
			\item HP\_TEMPBRGDRIVEEND
			\item HP\_TEMPBRGNONDRIVEEND
			\item HP\_TEMPBRGTHRUSTINBOARD
			\item HP\_TEMPBRGTHRUSTOUTBOARD
			\item HP\_TEMPDISCH1
			\item HP\_TEMPDISCH2
			\item HP\_TEMPLUBOIL
			\item HP\_TEMPLUBOILTANK
			\item HP\_TEMPSUCT1
			\item HP\_VIBAXIALDISP1
			\item HP\_VIBAXIALDISP2
			\item HP\_VIBDRIVEEND
			\item HP\_VIBDRIVEENDX
			\item HP\_VIBDRIVEENDY
			\item HP\_VIBNONDRIVEEND
			\item HP\_VIBNONDRIVEENDX
			\item HP\_VIBNONDRIVEENDY
			\item HP\_VOLDISCH
			\item HP\_VOLRATIO
			\item HP\_VOLSUCT
			\item LP\_HEADANTISURGE
			\item LP\_POWERSHAFT
			\item LP\_PRESSCLEANGAS
			\item LP\_PRESSDIFANTISURGE
			\item LP\_PRESSDIFSUCTSTRAINER
			\item LP\_PRESSDISCH
			\item LP\_PRESSSEALDRYGAS
			\item LP\_PRESSSEALLEAKPRIMARYDE1
			\item LP\_PRESSSEALLEAKPRIMARYDE2
			\item LP\_PRESSSEALLEAKPRIMARYNDE1
			\item LP\_PRESSSEALLEAKPRIMARYNDE2
			\item LP\_PRESSSUCT1
			\item LP\_PRESSSUCT2
			\item LP\_SPEED
			\item LP\_TEMPBRGDRIVEEND
			\item LP\_TEMPBRGNONDRIVEEND
			\item LP\_TEMPBRGTHRUSTINBOARD
			\item LP\_TEMPBRGTHRUSTOUTBOARD
			\item LP\_TEMPDISCH1
			\item LP\_TEMPDISCH2
			\item LP\_TEMPLUBOIL
			\item LP\_TEMPLUBOILTANK
			\item LP\_TEMPSUCT1
			\item LP\_VIBAXIALDISP1
			\item LP\_VIBAXIALDISP2
			\item LP\_VIBDRIVEEND
			\item LP\_VIBDRIVEENDX
			\item LP\_VIBDRIVEENDY
			\item LP\_VIBNONDRIVEEND
			\item LP\_VIBNONDRIVEENDX
			\item LP\_VIBNONDRIVEENDY
			\item LP\_VOLDISCH
			\item LP\_VOLRATIO
			\item LP\_VOLSUCT
			\item PT\_POWERSHAFT
			\item PT\_SPEED
			\item PT\_TEMPBRGDRIVEEND
			\item PT\_TEMPBRGNONDRIVEEND
			\item PT\_TEMPBRGTHRUST1
			\item PT\_TEMPBRGTHRUST3
			\item PT\_TEMPCOOLINGAIR1
			\item PT\_TEMPCOOLINGAIR2
			\item PT\_TEMPEXH
			\item PT\_TEMPLUBOIL
			\item PT\_TEMPLUBOILPTSUMP
			\item PT\_TEMPLUBOILTANK
			\item PT\_VIBAXIALDISP1
			\item PT\_VIBAXIALDISP2
			\item PT\_VIBBRGCASINGVEL
			\item PT\_VIBDRIVEEND
			\item PT\_VIBNONDRIVEEND
		}
	\end{enumerate}
\end{multicols}

\end{document}